\documentclass{article}
\usepackage{times}

\usepackage{amsmath,amsfonts,bm}

\def\eqref#1{equation~\ref{#1}}

\def\1{\bm{1}}

\DeclareMathAlphabet{\mathsfit}{\encodingdefault}{\sfdefault}{m}{sl}
\SetMathAlphabet{\mathsfit}{bold}{\encodingdefault}{\sfdefault}{bx}{n}

\usepackage{titling} 
\usepackage{amsmath,amssymb,amsthm}
\usepackage[colorlinks = true, linkcolor = black,  urlcolor  = blue, citecolor = black]{hyperref}
\usepackage[dvipsnames]{xcolor}
\usepackage{enumitem}
\usepackage{booktabs}
\usepackage{subcaption}
\usepackage{times}
\usepackage{graphicx, float}
\usepackage{mathtools}
\usepackage{natbib} 
\makeatletter
\newcommand{\papernotice}{}

\newcommand{\toptitlebar}{\hrule height 4pt\relax \vskip 0.25in \vskip -\parskip}
\newcommand{\bottomtitlebar}{\vskip 0.22in \vskip -\parskip \hrule height 1pt\relax \vskip 0.09in}

\renewcommand{\maketitle}{%
  \par
  \begingroup
    \renewcommand{\thefootnote}{\fnsymbol{footnote}}%
    \renewcommand{\@makefnmark}{\hbox to \z@{$^{\@thefnmark}$\hss}}%
    \thispagestyle{empty}%
    \toptitlebar
    \centering
    {\LARGE\bfseries \@title\par}
    \bottomtitlebar
    \begingroup
      \def\And{%
        \end{tabular}\hfil\linebreak[0]\hfil
        \begin{tabular}[t]{c}\bfseries\rule{\z@}{24pt}%
      }%
      \def\AND{%
        \end{tabular}\hfil\linebreak[4]\hfil
        \begin{tabular}[t]{c}\bfseries\rule{\z@}{24pt}%
      }%
      \begin{tabular}[t]{c}\bfseries\rule{\z@}{24pt}\@author\end{tabular}%
    \endgroup
    \vskip 0.3in\@minus 0.1in
    \@thanks
    \vfill
    \ifx\papernotice\@empty\else
      {\footnotesize \papernotice\par}
    \fi
  \endgroup
  \setcounter{footnote}{0}%
}
\makeatother

\title{A PDE-Informed Latent Diffusion Model for 2-m Temperature Downscaling}
\author{
  Paul Rosu\thanks{*Equal contribution}\\
  Department of Computer Science\\
  Duke University\\
  Durham, NC 27708 \\
  \texttt{paul.rosu@duke.edu} \\
  \And
  Muchang Bahng\footnotemark[1]\\
  Department of Computer Science\\
  Duke University\\
  Durham, NC 27708 \\
  \texttt{muchang.bahng@duke.edu} \\
  \And
  Erick Jiang\footnotemark[1]\\
  Department of Computer Science\\
  Duke University\\
  Durham, NC 27708 \\
  \texttt{erick.jiang@duke.edu} \\
  \And
  Rico Zhu\footnotemark[1]\\
  Department of Computer Science\\
  Duke University\\
  Durham, NC 27708 \\
  \texttt{rico.zhu@duke.edu} \\
  \And
  Vahid Tarokh \\
  Department of Electrical Engineering\\
  Duke University\\
  Durham, NC 27708 \\
  \texttt{vahid.tarokh@duke.edu} \\
}

\begin{document}
\maketitle

\begin{abstract}
    This work presents a physics-conditioned latent diffusion model tailored for dynamical downscaling of atmospheric data, with a focus on reconstructing high-resolution 2-m temperature fields. Building upon a pre-existing diffusion architecture and employing a residual formulation against a reference UNet, we integrate a partial differential equation (PDE) loss term into the model’s training objective. The PDE loss is computed in the full resolution (pixel) space by decoding the latent representation and is designed to enforce physical consistency through a finite-difference approximation of an effective advection-diffusion balance. Empirical observations indicate that conventional diffusion training already yields low PDE residuals, and we investigate how fine-tuning with this additional loss further regularizes the model and enhances the physical plausibility of the generated fields. The entirety of our codebase is available on Github\footnote{\href{https://github.com/paulrosu11/Physically-conditioned-latent-diffusion-model-for-temperature}{https://github.com/paulrosu11/Physically-conditioned-latent-diffusion-model-for-temperature}}, for future reference and development. 
\end{abstract}

\section{Introduction}

    Earth system models (ESMs) are critical for weather forecasting, tracking weather extremes, and supporting impact studies. In particular, numerical weather prediction (NWP) methods track surface and atmospheric data by dissecting the Earth's surface into grids, tracking variables of interest (e.g., temperature, wind speed, direction) as scalar/vector fields, and numerically solving partial differential equations (PDEs) to either physically interpolate into unknown regions or temporally evolve the model---a process known as reanalysis \cite{ritchie1995implementation, molteni1996ecmwf}. Historical reanalysis datasets such as ERA5, MERRA-2, and NCEP primarily consist of coarse-scale grid resolutions of $31 \times 31$ km to $500 \times 500$ km collected by weather stations, aircrafts, and meterological satellites \cite{dee2011era, merra2, ncep}. However, climate simulations at finer resolutions down to $2 \times 2$ km are critical for understanding short-term forecasting (nowcasting and medium-range forecasting) and predicting localized weather extremes described by highly resolved fields. 
    
    As manual collection of such high-resolution data on a global scale is too resource-intensive, global climate models (GCMs) perform downscaling to increase the resolution of surface data by employing two general types of techniques: dynamical and statistical downscaling \cite{balaji2022general, lam2023learning}. Dynamical downscaling uses regional climate models to extrapolate the effects recorded by large-scale models, but can be computationally prohibitive and largely rely on low-order functions to approximate hyperparameters, which have been shown to introduce bias \cite{allen2002}. On the other hand, statistical downscaling uses models such as multilinear regression and ensemble techniques to model relationships between global and local conditions. 
    
    In particular, deep learning approaches based on CNNs, GANs, and diffusion models have shown promise in bridging the gap between coarse global outputs and fine-scale atmospheric features \cite{saoulis2025diffusion, tarokh}. However, while these models often optimize for statistical similarity, they rarely incorporate explicit physical constraints. Many physical phenomena are governed by PDEs, such as the Navier–Stokes equations in fluid dynamics, the heat equation in thermal analysis, and Maxwell’s equations in electromagnetism. Physics-informed neural networks (PINNs) embed knowledge of physical laws by incorporating these governing equations directly into the training process as part of the loss function. Consider a PDE defined on a domain $\Omega$ as $\mathcal{N}[u(x,t)] = 0$, where $\mathcal{N}$ is a differential operator capturing the underlying physics, and $u(x,t)$ is the solution function. In PINNs, the solution is approximated by a neural network $u(x,t;\theta)$ with parameters $\theta$. Approximations to differentiation are employed to compute the necessary derivatives of $u(x,t;\theta)$ with respect to $x$ and $t$. For example, if $\mathcal{N}$ involves the time derivative $u_t$ and spatial derivatives $u_x$ or $u_{xx}$, these can be calculated as
    \begin{equation*}
        u_t(x,t;\theta) = \frac{\partial u(x,t;\theta)}{\partial t}, \quad u_x(x,t;\theta) = \frac{\partial u(x,t;\theta)}{\partial x}, \quad u_{xx}(x,t;\theta) = \frac{\partial^2 u(x,t;\theta)}{\partial x^2}.
    \end{equation*}
    The PDE residual is then defined by substituting the network approximation into the differential operator:
    \begin{equation*}
        r(x,t;\theta) = \mathcal{N}[u(x,t;\theta)].
    \end{equation*}
    To ensure that the network output adheres to the physics described by the PDE, a loss term is constructed as the mean squared error of the residual over a set of $N$ collocation points $\{(x_i,t_i)\}_{i=1}^{N}$:
    \begin{equation*}
        \mathcal{L}_{\text{PDE}}(\theta) = \frac{1}{N}\sum_{i=1}^{N} \left| \mathcal{N}[u(x_i,t_i;\theta)] \right|^2.
    \end{equation*}
    \noindent In addition to the PDE residual loss, PINNs incorporate initial and boundary condition constraints. For instance, if the initial condition is given by $u(x,0) = u_{\text{IC}}(x)$ for $M$ sample points $\{x_j\}_{j=1}^{M}$, the corresponding loss term is defined as
    \begin{equation*}
        \mathcal{L}_{\text{IC}}(\theta) = \frac{1}{M}\sum_{j=1}^{M} \left| u(x_j,0;\theta) - u_{\text{IC}}(x_j) \right|^2.
    \end{equation*}
    Similarly, a boundary condition loss $L_{\text{BC}}(\theta)$ is computed over points on the boundary $\partial\Omega$. We then take a linear combination of the PDE losses to produce the final composite loss. 
    
    To integrate the best of both worlds, we propose a latent diffusion framework that is augmented with an explicit PDE loss on 2-m temperature (temperature 2 meters above the surface). This PDE loss not only promotes physical consistency but also serves as a regularization term that mitigates unphysical artifacts in the high-resolution outputs. The remainder of this paper is organized as follows. In Section~\ref{sec:related}, we briefly review related downscaling techniques and generative models. Section~\ref{sec:method} details the methodology, with a strong focus on the design and numerical implementation of the PDE loss. Section~\ref{sec:experiments} describes the experimental setup and training details. In Section~\ref{sec:results}, we provide an overview of the evaluation metrics and qualitative results, and Section~\ref{sec:discussion} concludes with discussion and limitations.

\section{Related Work}
\label{sec:related}

    \paragraph{Climate Modeling with Deep Neural Nets} Several architectures have been employed in predicting atmospheric variables in both global and local regions. 
    In particular, convolutional neural networks (CNNs), which are tailored for spatial grid-like data and capture local advection within neighboring pixels, have been trained to postprocess GCM 2-m temperature forecasts and downscale precipitation levels on a finer horizontal resolution \cite{rasp2018, rodrigues2018}. To predict atmospheric pressure at a future point in time, forecasting models such as the Deep Learning Weather Prediction (DLWP) extend existing encoder-decoder architectures by incorporating convolutional LSTM layers to outperform their non-recurrent counterparts \cite{canmach}. Other models such as the 3D Earth Specific Transformer (3DEST) consist of a collection of trained encoder-decoder models that each predict atmospheric pressure and wind speeds over a fixed time into the future (e.g. models FM1, FM3, FM6, and FM24 predict wind speeds up to 1, 3, 6, and 24 hours into the future), with inference done in a few minutes compared to several hours for NWP \cite{3dneural}. Atmospheric pressure predictors have been directly applicable for detecting cyclones, which occur in regions with low sea-level pressure. The increase in accuracy and performance with deep learning spurred the creation of more datasets tailored for deep learning \cite{sevir}. 
    
    \paragraph{Climate Diffusion Models} Diffusion models have two principal applications in climate science: (1) forecasting, and (2) data downscaling. Climate forecasting models typically involve modeling the conditional distribution $p(x_{t+1:t+h}\mid x_{t-l+1:t})$ that predicts $h$ frames in the future given $l$ snapshots from the past. Toward this problem, DYffusion \cite{ruhling2023dyffusion} proposes a novel coupling mechanism that directly mixes the temporal process of the time-series data directly with the diffusion process; LDCast \cite{leinonen2023ldcast} uses Adaptive Fourier Neural Operators \cite{guibas2021adaptive} as the encoder to learn latent embeddings spatiotemporal climate data in the Fourier domain, and then uses a LDM over the latent space to generate forecasts.\\
    
    \noindent Our work focuses on data downscaling, which is the task of making fine-grained predictions about localized weather events using global climate information. Diffusion models come in once again as an effective, general generative method for learning both unconditional and conditional distributions \cite{schmidt2024spatiotemporally, watt2024generative}. DiffScaler \cite{tomasi2024can} uses a latent diffusion model (LDM) for learning the conditional distribution between low-resolution spatiotemporal data mixed with high-resolution static information in order to make predictions about local wind speeds and temperature conditions; \cite{srivastava2024precipitation} performs downscaling by decomposing the process into two main components: a deterministic downscaling module that produces an initial high‐resolution estimate, and a stochastic diffusion module that refines this estimate by predicting the high-frequency details as a residual.

\section{Methodology}
\label{sec:method}

    Our proposed framework builds upon a latent diffusion model (LDM) originally developed for downscaling tasks \cite{rombach2022high, leinonen2023ldcast}. In our variant, we fine-tune the final 200 million parameters of the LDM from Tomasi et al. to incorporate an explicit PDE loss \cite{tomasi2024can}. This section details both the overall architecture and, in particular, the PDE loss implementation.

\subsection{Problem Formulation}
\label{subsec:formulation}

\begin{figure}
        \centering
        \includegraphics[width=0.95\textwidth]{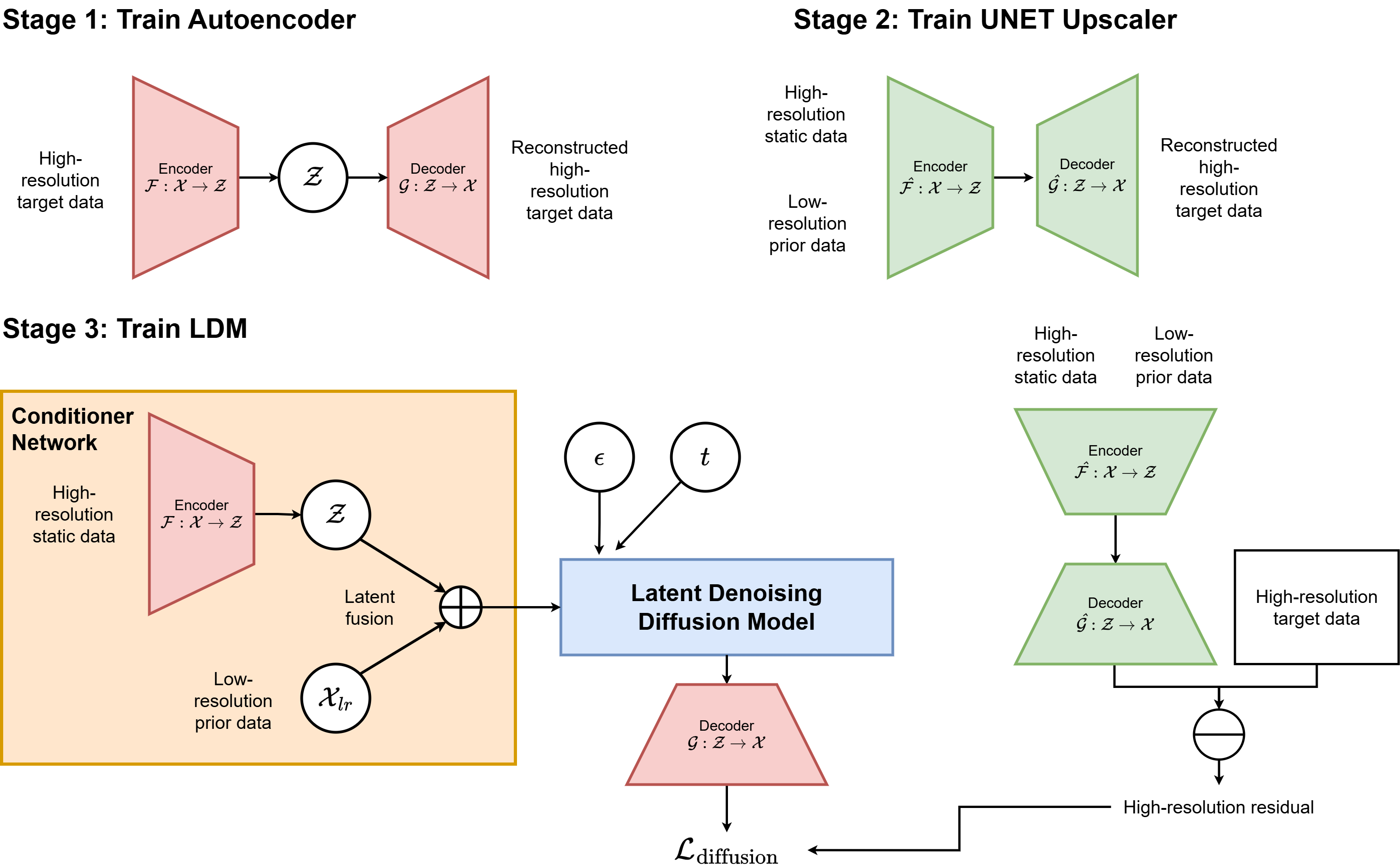}
        \caption{Training pipeline for our model. We train in two stages, where we first pretrain an autoencoder and upscaler, and then use the learned encoder in the conditioner network for the LDM, as well as the upscaler to compute the residual.}
        \label{fig:training-LDM}
    \end{figure}

    \noindent Dynamic downscaling describes the upscaling of low-resolution climate data as outputted from a General Circulation Model (GCM), via a more specialized Regional Climate Model (RCM). The goal of our work is to investigate whether we can use diffusion models as an efficient and effective approximator for RCMs in the downscaling process. We formulate this process as a conditional generative process, where we condition on low-resolution information $X_{lr}$ in the denoiser to extract a high-resolution image $X_{hr}$. If we fix the dimensionality of the images, we can consider the low-resolution features as a prior in that our model parametrizes the distribution $p(X_{hr} \mid X_{lr})$. In addition to low-resolution GCM data, we also consider some static high-resolution features such as features extracted from a Digital Elevation Model (DEM), land cover categories, and latitude information. We denote this as $X_{\text{static}\, hr}$, and rewrite the target as
    \begin{equation*}
        p(X_{hr} \mid X_{lr}, X_{\text{static}\, hr})
    \end{equation*}
    One assumption we make is that $X_{\text{static}\, hr} \sim X_{hr}$, so that we can treat both as belonging to the same space. \cite{mardani2025residual, tomasi2024can} have demonstrated empirically that diffusion models work well in complementing existing upscaling models. Instead of generating high-resolution images directly, we first pretrain a upscaling model (e.g., UNet) and then use the diffusion model to predict the difference between the upscaled image and the ground-truth. The diffusion model is used as a corrector model instead. A diagram of the entire training pipeline can be seen in Figure \ref{fig:training-LDM}.

\subsection{Latent Diffusion Architecture}
\label{subsec:architecture}

    The core idea behind diffusion models is to gradually corrupt data with noise and then learn a reverse (denoising) process to reconstruct the original data \cite{sohl2015deep}. The model described in this section closely adheres to existing architectures for the exact same problem \cite{tomasi2024can}. In our model, the forward process is applied in a latent space obtained via a Variational AutoEncoder (VAE), which we fit by minimizing the L2-reconstruction loss between ground-truth high-resolution data and the reconstructions. So we first pretrain both a VAE and UNet for compression and upscaling, respectively. The VAE gives us a mappings to and from the high-resolution target space to the latent diffusion space, and the UNet is fitted to upscale images in-taking low-resolution data and high-resolution static data (same as the diffusion model).
    
    \noindent For the latent diffusion model, we first  apply the pretrained VAE encoder $F$ directly on the static high-resolution inputs $X_{hr}$ to yield the low-dimensional latent representations. For the low-resolution conditioning variables $X_{lr}$, we project the inputs into the same low-dimensional space as the VAE embedding space via a convolutional layer. To get a more complete characterization of the latent space, we fuse both the low and high-resolution embeddings by concatenating both and then feeding the resulting product vector into a linear network. We denote this part of the model as the conditioner:
    \begin{equation*}
        \begin{split}
            Z &= \text{Linear}\left( F(X_{\mathrm{static}\, hr}) \,\oplus\, \text{ConvProj}(X_{lr}) \right)
        \end{split}
    \end{equation*}
    where $\oplus$ denotes concatenation.\\
    
    \noindent During training, the denoiser network $\epsilon_\theta$ (with UNet-like structure) is conditioned on low-resolution predictors and high-resolution static information via the conditioning network. We first pass the low-resolution and high-resolution static data into both the pretrained upscaler and subtract that from the ground-truth high-resolution image to obtain the residual. In a standard setting, the model minimizes the denoising diffusion objective $\mathcal{L}_{\text{diffusion}}$:
    \begin{equation*}
        \mathcal{L}_{\text{diffusion}} = \mathbb{E}_{Z_0\sim q(Z_0),\,\epsilon\sim\mathcal{N}(0,I), \,t\sim\mathcal{U}(0, T)}\left[ \| \epsilon - \epsilon_\theta(\sqrt{\bar{\alpha}_t}\,Z_0 + \sqrt{1 - \bar{\alpha}_t}\, \epsilon, \, t) \|^2 \right].
    \end{equation*}
    where $\mathcal{U}$ is the uniform distribution, $\alpha_t$ is the noise scheduler, the diffusion output is compared against the residual. We extend this objective by adding a PDE loss term to encourage physicality:
    \begin{equation*}
      \mathcal{L} = \mathcal{L}_{\text{diffusion}} + \lambda_{\text{PDE}}\,\mathcal{L}_{\text{PDE}},
    \end{equation*}
    where $\lambda_{\text{PDE}}$ is a hyperparameter balancing the two objectives.

\subsection{Physics-Constrained PDE Loss Formulation}
\label{subsec:flux_loss}

The central motivation behind our PDE-informed latent diffusion approach lies in enforcing physically consistent structures on generated high-resolution temperature fields. Atmospheric temperature dynamics, particularly near Earth's surface, typically reflect a balance between advective transport, which moves thermal energy along flow patterns, and diffusive processes, which smooth temperature gradients \cite{book}. Although in true atmospheric physics, the advection term explicitly involves wind vector fields, our architecture intentionally excludes wind data for fair comparison with existing downscaling models that only use snapshots of temperature fields during training and inference. Therefore, we approximate advection implicitly through the  temperature gradients themselves. Physically, this is motivated by observing that large-scale temperature patterns inherently reflect dominant transport directions and intensities. 

\paragraph{Mathematical Framework and Multi-Scale Representation}
Given the coarse-scale temperature field \(T_c: \Omega \subset \mathbb{R}^2 \to \mathbb{R}\), our goal is to predict a high-resolution temperature field \(T_f\). We define the AI model as an operator:
\[
\mathcal{M}_{\mathrm{LR}\to\mathrm{HR}}: L^2(\Omega) \to L^2(\Omega), \quad T_f = \mathcal{M}_{\mathrm{LR}\to\mathrm{HR}}(T_c).
\]
To handle boundary value ambiguities inherent to individual coarse cells, we introduce an additional, even coarser partition termed the \emph{supergrid}. Each supergrid cell \(\Omega_\Gamma\subset\Omega\) aggregates multiple coarse cells, allowing us to define clear boundary conditions for computing physically meaningful fluxes at both coarse and fine scales.

\paragraph{Advection–Diffusion Flux Balance}
For each supergrid cell \(\Omega_\Gamma\) with boundary \(\partial\Omega_\Gamma\), conservation of thermal energy allows us to define:
\[
\Phi_{\Gamma} \coloneq \oint_{\partial \Omega_\Gamma} \mathbf{F}\cdot\mathbf{n}\,dS,
\]
where the total flux \(\mathbf{F}\) is decomposed into advective and diffusive components:
\[
\mathbf{F}(x) = \mathbf{u}(x)T(x) - D\nabla T(x).
\]
Due to the absence of explicit wind data, we approximate the unknown velocity field \(\mathbf{u}(x)\) using the gradient direction of temperature, justified by the empirical correlation between temperature gradients and the dominant thermal transport direction. Specifically, we introduce a stabilized gradient unit vector:
\[
\widehat{g}(x) = \frac{\nabla T(x)}{\|\nabla T(x)\| + \varepsilon},
\]
where \(\varepsilon > 0\) ensures numerical stability.

\paragraph{Flux Definitions on the Supergrid}
The effective advective and diffusive fluxes across each supergrid boundary are computed via finite differences. The coarse-grid temperature derivatives are approximated as:
\begin{align*}
\partial_x T_c(i,j) &\approx \frac{T_c(i,j+1) - T_c(i,j-1)}{2\Delta x}, \\
\partial_y T_c(i,j) &\approx \frac{T_c(i+1,j) - T_c(i-1,j)}{2\Delta y},
\end{align*}
leading to the gradient magnitude approximation:
\[
\|\nabla T_c(i,j)\| \approx \sqrt{\left(\frac{T_c(i,j+1)-T_c(i,j-1)}{2\Delta x}\right)^2 + \left(\frac{T_c(i+1,j)-T_c(i-1,j)}{2\Delta y}\right)^2}.
\]

Thus, we define the advective and diffusive fluxes for each supergrid cell boundary as:
\begin{align*}
\Phi_{\mathrm{adv},\Gamma} &\propto \frac{1}{|\partial \Omega_\Gamma|}\sum_{x\in\partial\Omega_\Gamma} T(x)(\widehat{g}(x)\cdot\mathbf{n}(x)), \\
\Phi_{\mathrm{diff},\Gamma} &\propto \frac{1}{|\partial \Omega_\Gamma|}\sum_{x\in\partial\Omega_\Gamma} |\nabla T(x)|,
\end{align*}

where these fluxes are computed at both coarse and fine scales after appropriate aggregation. The dimensionless effective flux ratio (analogous to a Péclet number) is:
\[
R_{\mathrm{eff}} = \frac{\Phi_{\mathrm{adv},\Gamma}}{\Phi_{\mathrm{diff},\Gamma} + \varepsilon}.
\]
This ratio, computed from coarse (\(R^{(c)}_{\mathrm{eff}}\)) and fine-scale predictions (\(R^{(f)}_{\mathrm{eff}}\)), quantifies the balance between advective and diffusive processes.

\paragraph{Multi-Scale Loss Function}
To enforce physical consistency, we introduce the PDE loss based on flux ratio discrepancies:
\[
\mathcal{L}_{\mathrm{PDE}} = \frac{1}{N}\sum_{\Gamma} \left(R^{(f)}_{\mathrm{eff},\Gamma} - R^{(c)}_{\mathrm{eff},\Gamma}\right)^2,
\]
where \(N\) is the number of supergrid cells. This loss explicitly encourages the fine-scale model outputs to maintain the same advective–diffusive flux balance observed at the coarse level.

\paragraph{Discussion on Computational Complexity}
Computational efficiency is crucial for integrating PDE constraints in model training. Our finite-difference approach allows rapid, parallelized computations using efficient matrix multiplications optimized on GPUs. The introduction of supergrid cells introduces a complexity trade-off: larger cells reduce computational overhead but may oversimplify boundary definitions, while smaller cells improve physical accuracy at increased computational cost. For maximal efficiency, we choose supergrid cell sizes based on the greatest common divisor of coarse grid dimensions along the x and y axes, maximizing speed. This choice is computationally pragmatic but remains a hyperparameter that warrants experimental tuning.

Additionally, by performing PDE loss computations in decoded pixel space, we bridge the computational efficiency of latent-space modeling with the interpretability and physical fidelity required in the high-resolution output domain. Ultimately, this formulation achieves a robust compromise between physical rigor and computational practicality, enhancing both the interpretability and reliability of our latent diffusion model for atmospheric downscaling.

\section{Experimental Setup}
\label{sec:experiments}

Our experimental framework builds upon the work of \cite{leinonen2023ldcast} and \cite{tomasi2024can}, focusing solely on 2-m temperature downscaling. We use ERA5 reanalysis data \cite{era5data} as low-resolution input and high-resolution COSMO-CLM \cite{rockel2008performance} simulations as target data for 2-m temperature. The dataset centered on Italy that we used was compiled and made publicly available by a previous group of researchers \cite{raffa2021cosmo}. For a list of the specific features we extracted to train our model and for inference, refer to Appendix \ref{app:features}. Our model was trained and tested on the full set of data timestamps, with a 0.7/0.15/0.15 split between training, validation, and testing sets. The dataset is easily accessible through a Zenodo reposoritry with download scripts provided by both us and Tomasi et al. \cite{tomasi2024can}. 

The latent diffusion model comprises a VAE, a UNet-based denoiser, and a conditioning network. The model is trained in a residual setting, where the output of a pre-trained reference UNet is subtracted before encoding and added back after decoding. Fine-tuning was performed on four NVIDIA A5000 GPUs over a total of 12 GPU hours. Inference requires a single A5000 GPU. All GPUs were accessed through jobs queued on the Duke Computer Science Department's cluster. Due to GPU memory constraints, we fine-tuned only the last 200 million parameters of the LDM. The hyperparameter $\lambda_{\text{PDE}}$ (see Section  \ref{subsec:architecture}) was chosen based on preliminary experiments to balance the diffusion and PDE losses.


\section{Results}
\label{sec:results}

We compare our \textbf{PDE–constrained latent diffusion model} (\textbf{LDM\textsubscript{PDE}}) with quadratic interpolation and the UNet, GAN, and the state‑of‑the‑art residual latent diffusion model (\textbf{LDM\textsubscript{res}}) provided by Tomasi et al \cite{tomasi2024can}. We refer to Tomasi et al. for the details of these benchmark models as the main interest of our work is the virtue of PDE regulation in the loss for difusion models. Evaluation is performed on the held‑out set of  COSMO‑CLM target scenes using 
\emph{four statistical} and \emph{two physics‑aware} metrics, thereby considering pixel‑wise accuracy as well as physical plausibility.

\subsection{Metrics}
\label{subsec:metrics}

\paragraph{Statistical metrics}  Let $T_{f}^{(i)}$ and $T_{t}^{(i)}$ denote the predicted and ground‑truth temperatures at pixel $i\!=\!1,\dots ,N$.  Define the mean of field $\mathbf T$ as $\bar T\!=\!\tfrac1N\sum_iT^{(i)}$. Then the formulas for the statistical metrics we apply are
\begin{align}
  \mathrm{RMSE} &= \sqrt{\frac1N\sum_{i=1}^{N}\bigl(T_{f}^{(i)}-T_{t}^{(i)}\bigr)^{2}}   \tag{RMSE} \\
  R^{2} &= 1-\frac{\sum_i\bigl(T_{f}^{(i)}-T_{t}^{(i)}\bigr)^{2}}{\sum_i\bigl(T_{t}^{(i)}-\bar T_{t}\bigr)^{2}} \tag{R2} \\
  \mathrm{PCC} &= \frac{\sum_i\bigl(T_{f}^{(i)}-\bar T_{f}\bigr)\bigl(T_{t}^{(i)}-\bar T_{t}\bigr)}{\sqrt{\sum_i\bigl(T_{f}^{(i)}-\bar T_{f}\bigr)^{2}}\,\sqrt{\sum_i\bigl(T_{t}^{(i)}-\bar T_{t}\bigr)^{2}}}   \tag{PCC} \\
  \mathrm{Bias} &= \frac1N\sum_{i=1}^{N}\bigl(T_{f}^{(i)}-T_{t}^{(i)}\bigr)   \tag{BIAS}
\end{align}

\paragraph{Physics‑aware metrics}
\begin{enumerate}[leftmargin=*, nosep]
  \item \textbf{Flux‑ratio loss} \(\displaystyle \mathcal L_{\mathrm{flux}}=\tfrac1N\sum_{\Gamma}(R_{\!\mathrm{eff},\Gamma}^{(f)}-R_{\!\mathrm{eff},\Gamma}^{(c)})^{2}\) (See Section~\ref{subsec:flux_loss}).
  \item \textbf{Spectral‑slope difference loss} \(\mathcal L_{\mathrm{spec}} = |\alpha_f - \alpha_c| \) (defined below in Section \ref{subsec:ralsd}).
\end{enumerate} 

\subsection{Radially Averaged Log–Spectral Density (RALSD)}
\label{subsec:ralsd}

To evaluate how physically accurate each model's predictions are, we employ the \emph{radially averaged log–spectral density (RALSD)}, as introduced in previous climate modeling studies \citep{Harris_2022,Ravuri_2021}.

Given a predicted high-resolution temperature  field $T_f(x,y)$, defined on a uniform spatial grid, we first compute its two-dimensional discrete Fourier transform (DFT):
\[ \widehat{T}_f(k_x, k_y) 
  = \sum_{x,y} T_f(x,y)\,\exp\left[-2\pi i \left(\frac{k_x x}{H} + \frac{k_y y}{W}\right)\right], \]
where $H$ and $W$ denote grid dimensions. The associated power spectral density (PSD) is defined by:
\[  P_f(k_x, k_y) = |\widehat{T}_f(k_x, k_y)|^2. \]

Next, we radially average the PSD by grouping Fourier coefficients into concentric annuli defined by the radial wavenumber $k = \sqrt{k_x^2 + k_y^2}$:
\[ \Psi_f(k_i) = \frac{1}{|\mathcal{K}_i|}\sum_{(k_x,k_y)\in\mathcal{K}_i} P_f(k_x,k_y), \]
where $\mathcal{K}_i$ represents the set of spectral coefficients within the $i^\text{th}$ radial bin.

To characterize the cascade of variance across spatial scales, we plot $\Psi_f(k_i)$ against $k_i$ in log-log space, obtaining the RALSD curve \cite{RampalRALSD, Harris_2022}. By performing a linear least-squares fit over the inertial range, we extract the spectral slope $\alpha_f$. Repeating this procedure on the coarse-grid, up-sampled reference field yields the baseline slope $\alpha_c$. Finally, the spectral-slope difference loss is defined as:
\[   \mathcal{L}_{\mathrm{spec}} = |\alpha_f - \alpha_c|, \]
with lower values being better. This approach is grounded in the observation that atmospheric fields often exhibit approximate power-law behavior in their spatial spectra. For an extensive discussion on the interpretation and implications of linear fits in log–log spectral analyses, refer to Lovejoy's 2023 in depth review article \citep{npg-30-311-2023}.

\subsection{Quantitative results}
\label{subsec:quantitative}

\begin{figure}[H]
  \centering
  \includegraphics[width=1\linewidth]{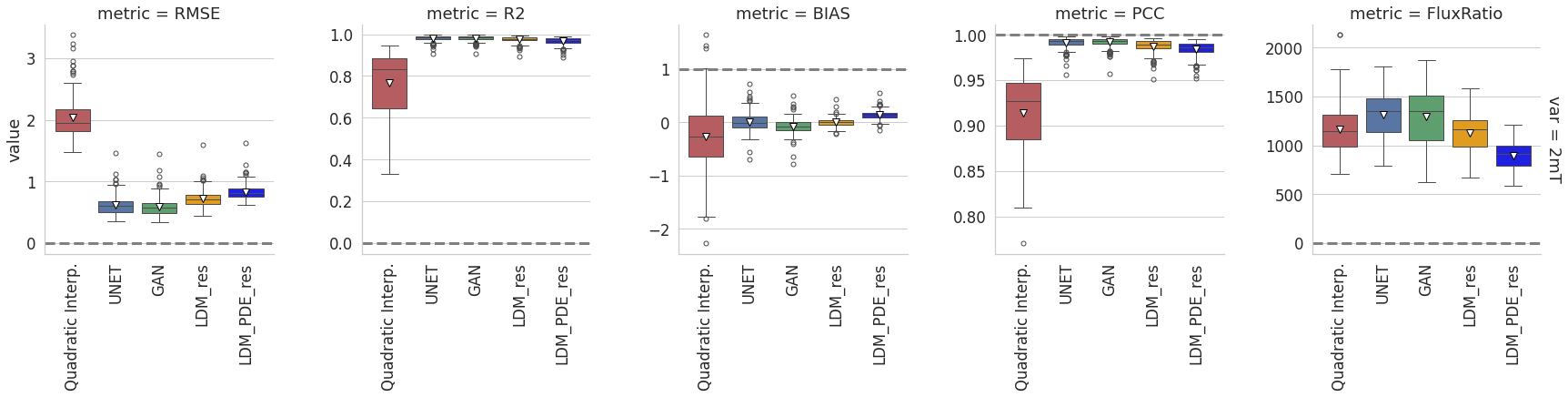}
  \caption{Comparison across RMSE, $R^{2}$, PCC, Bias, and flux‑ratio loss ($R_{\mathrm{eff}}$).  Lower bars denote better performance except for $R^{2}$ and PCC (higher is better).  The PDE‑constrained model achieves the best physics‑aware score while remaining competitive statistically.}
  \label{fig:results}
\end{figure}

\begin{figure}[H]
  \centering
  \includegraphics[width= .7\linewidth]{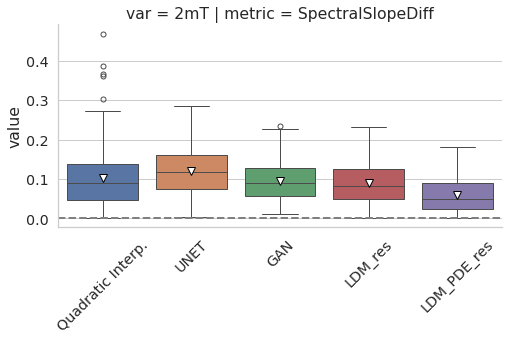}
  \caption{Spectral‑slope loss $\mathcal L_{\mathrm{spec}}$ over the test-set.  Lower values indicate closer agreement with the COSMO‑CLM RALSD slope.}
  \label{fig:spectral_box}
\end{figure}

Figure~\ref{fig:results} compares all models across the statistical metrics and the flux‑ratio metric.  Despite slightly higher RMSE and lower $R^{2}$, both \textbf{LDM\textsubscript{res}} and \textbf{LDM\textsubscript{PDE}} excel on the physics metric $\mathcal L_{\mathrm{flux}}$, indicating that pixel‑wise scores alone do not capture physical fidelity, further validated by Figure-\ref{fig:spectral_box}

Figure~\ref{fig:spectral_box} reports $\mathcal L_{\mathrm{spec}}$.  \textbf{LDM\textsubscript{PDE}} shows both the lowest median and the tightest interquartile range, confirming superior reproduction of the target energy cascade.

\subsection{Qualitative evaluation}
\label{subsec:qualitative}

\begin{figure}[H]
        \centering
        \includegraphics[width=0.9\linewidth]{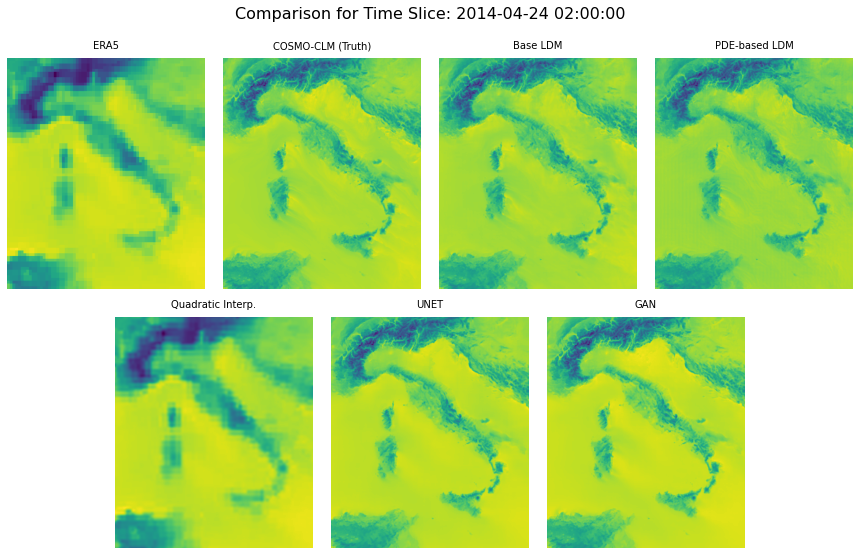}
        \caption{Visually, the UNet and GAN based models fail to capture the fine whisps of temperature fluctuations occurring over the Tyrrhenian Sea as shown in the ground truth (COSMO-CLM). The base LDM and our improved PDE-based LDM does capture these details. }
        \label{fig:comparison3}
    \end{figure}

As illustrated in Figure~\ref{fig:comparison3}, quadratic interpolation, UNet, and GAN smooth away jet‑like filaments evident in the COSMO-CLM ground truth.  \textbf{LDM\textsubscript{res}} recovers these structures but occasionally introduces speckle.  In contrast, \textbf{LDM\textsubscript{PDE}} retains fine detail while suppressing artefacts, producing notably smoother and more coherent temperature gradients, particularly in regions with complex heat flow dynamics. The PDE loss successfully mitigates discontinuities that are present in base LDM outputs, leading to improved visual correspondence with the ground-truth COSMO-CLM reference data. These visual improvements highlight the utility of enforcing explicit physical constraints. To allow further comparison, additional figures are included in Appendix \ref{app:figs} and the inference code for all the models are provided in the \href{https://github.com/paulrosu11/Physically-conditioned-latent-diffusion-model-for-temperature}{GitHub}. 

\section{Discussion, Limitations, and  Conclusion}
\label{sec:discussion}
\subsection{Limitations}
\label{subsec:limitations}

Our results, while promising, come with several caveats that warrant attention. First, in order to directly compare with existing work, we train the model on a single prognostic field---2-m temperature, ignoring the tight coupling between temperature, humidity, wind, and other state variables. This single-variable design forces us to approximate advective fluxes with latent temperature gradients rather than explicit wind vectors, a simplification that can misrepresent dynamics where winds are strong but thermal contrasts are weak. Second, similarly constrained by comparability, training and evaluation are conducted on a single regional dataset (Italy, ERA5 → COSMO-CLM). Thus, the tuned hyperparameter \(\lambda_{\text{PDE}}\) and the overall performance may not generalize to regions with significantly different orography, land–sea contrasts, or grid spacings. Broader intercomparisons across climates and resolutions are necessary to confirm robustness. Third, due to memory constraints, we fine-tune only the final 200 million parameters of the pre-trained latent diffusion model, leaving earlier layers, and any inherent statistical biases, unchanged; full end-to-end optimization could reveal additional improvements or new failure modes. Finally, we report deterministic scores and single-sample visuals only; while diffusion models naturally support ensemble generation, we have not yet assessed their ensemble spread, calibration, or reliability.

\subsection{Conclusion}

In this study, we introduced a PDE-constrained latent diffusion model for downscaling 2-m temperature data. By integrating a finite-difference-based PDE residual loss—computed efficiently in decoded pixel space—into the training objective, we enforce greater physical consistency in the high-resolution outputs without significantly compromising standard statistical metrics. A key innovation is our parameter-fitting strategy, which dynamically estimates effective diffusivity and advection parameters directly from coarse-scale inputs during training. While the PDE loss introduces modest additional computational overhead, inference remains competitive with existing state-of-the-art latent diffusion models and considerably faster than numerical weather methods such as COSMO-CLM \cite{tomasi2024can}.

\bibliographystyle{plain}
\bibliography{references}

\begin{thebibliography}{10}

\bibitem{allen2002}
Myles Allen, Jamie Kettleborough, and David Stainforth.
\newblock Model error in weather and climate forecasting.
\newblock {\em Proceedings of The IEEE - PIEEE}, 9780521848824, 01 2002.

\bibitem{tarokh}
Ahmed Aloui, Ali Hasan, Juncheng Dong, Zihao Wu, and Vahid Tarokh.
\newblock Score-based metropolis-hastings algorithms.
\newblock {\em CoRR}, abs/2501.00467, 2025.

\bibitem{balaji2022general}
V~Balaji, Fleur Couvreux, Julie Deshayes, Jacques Gautrais, Fr{\'e}d{\'e}ric Hourdin, and Catherine Rio.
\newblock Are general circulation models obsolete?
\newblock {\em Proceedings of the National Academy of Sciences}, 119(47):e2202075119, 2022.

\bibitem{3dneural}
Kaifeng Bi, Lingxi Xie, Hengheng Zhang, Xin Chen, Xiaotao Gu, and Qi~Tian.
\newblock Accurate medium-range global weather forecasting with 3d neural networks.
\newblock {\em Nature}, 619(7970):533--538, 2023.

\bibitem{dee2011era}
Dick~P Dee, S~M Uppala, Adrian~J Simmons, Paul Berrisford, Paul Poli, Shinya Kobayashi, U~Andrae, M~A Balmaseda, G~Balsamo, P~Bauer, P~Bechtold, A~C~M Beljaars, L~van~de Berg, J~Bidlot, N~Bormann, C~Delsol, R~Dragani, M~Fuentes, A~J Geer, L~Haimberger, S~B Healy, H~Hersbach, E~V H{\'o}lm, L~Isaksen, Per K{\aa}llberg, M~K{\"o}hler, M~Matricardi, A~P McNally, B~M Monge-Sanz, J-J Morcrette, B-K Park, C~Peubey, P~de~Rosnay, C~Tavolato, J-N Th{\'e}paut, and F~Vitart.
\newblock The era-interim reanalysis: Configuration and performance of the data assimilation system.
\newblock {\em Quarterly Journal of the Royal Meteorological Society}, 137(656):553--597, 2011.

\bibitem{merra2}
Ronald Gelaro, Will McCarty, Max~J. Suárez, Ricardo Todling, Andrea Molod, Lawrence Takacs, Cynthia~A. Randles, Anton Darmenov, Michael~G. Bosilovich, Rolf Reichle, Krzysztof Wargan, Lawrence Coy, Richard Cullather, Clara Draper, Santha Akella, Virginie Buchard, Austin Conaty, Arlindo~M. da~Silva, Wei Gu, Gi-Kong Kim, Randal Koster, Robert Lucchesi, Dagmar Merkova, Jon~Eric Nielsen, Gary Partyka, Steven Pawson, William Putman, Michele Rienecker, Siegfried~D. Schubert, Meta Sienkiewicz, and Bin Zhao.
\newblock The modern-era retrospective analysis for research and applications, version 2 (merra-2).
\newblock {\em Journal of Climate}, 30(14):5419 -- 5454, 2017.

\bibitem{guibas2021adaptive}
John Guibas, Morteza Mardani, Zongyi Li, Andrew Tao, Anima Anandkumar, and Bryan Catanzaro.
\newblock Adaptive fourier neural operators: Efficient token mixers for transformers.
\newblock {\em arXiv preprint arXiv:2111.13587}, 2021.

\bibitem{Harris_2022}
Lucy Harris, Andrew T.~T. McRae, Matthew Chantry, Peter~D. Dueben, and Tim~N. Palmer.
\newblock A generative deep learning approach to stochastic downscaling of precipitation forecasts.
\newblock {\em Journal of Advances in Modeling Earth Systems}, 14(10), October 2022.

\bibitem{era5data}
Hans Hersbach, Bill Bell, Paul Berrisford, Gionata Biavati, Andr\'as Hor\'anyi, Joaqu\'in Mu\~noz Sabater, Julien Nicolas, Carole Peubey, Raluca Radu, Iryna Rozum, Dinand Schepers, Adrian Simmons, Cornel Soci, Dick Dee, and Jean-No\"el Th\'epaut.
\newblock Era5 hourly data on single levels from 1940 to present. copernicus climate change service (c3s) climate data store (cds), 2023.

\bibitem{ncep}
E.~Kalnay, M.~Kanamitsu, R.~Kistler, W.~Collins, D.~Deaven, L.~Gandin, M.~Iredell, S.~Saha, G.~White, J.~Woollen, Y.~Zhu, M.~Chelliah, W.~Ebisuzaki, W.~Higgins, J.~Janowiak, K.~C. Mo, C.~Ropelewski, J.~Wang, A.~Leetmaa, R.~Reynolds, Roy Jenne, and Dennis Joseph.
\newblock The ncep/ncar 40-year reanalysis project.
\newblock {\em Bulletin of the American Meteorological Society}, 77(3):437 -- 472, 1996.

\bibitem{lam2023learning}
Remi Lam, Alvaro Sanchez-Gonzalez, Matthew Willson, Peter Wirnsberger, Meire Fortunato, Ferran Alet, Suman Ravuri, Timo Ewalds, Zach Eaton-Rosen, Weihua Hu, et~al.
\newblock Learning skillful medium-range global weather forecasting.
\newblock {\em Science}, 382(6677):1416--1421, 2023.

\bibitem{leinonen2023ldcast}
Jussi Leinonen, Ulrich Hamann, Daniele Nerini, Urs Germann, and Gabriele Franch.
\newblock Latent diffusion models for generative precipitation nowcasting with accurate uncertainty quantification, 2023.

\bibitem{npg-30-311-2023}
S.~Lovejoy.
\newblock Review article: Scaling, dynamical regimes, and stratification. how long does weather last? how big is a cloud?
\newblock {\em Nonlinear Processes in Geophysics}, 30(3):311--374, 2023.

\bibitem{mardani2025residual}
Morteza Mardani, Noah Brenowitz, Yair Cohen, Jaideep Pathak, Chieh-Yu Chen, Cheng-Chin Liu, Arash Vahdat, Mohammad~Amin Nabian, Tao Ge, Akshay Subramaniam, et~al.
\newblock Residual corrective diffusion modeling for km-scale atmospheric downscaling.
\newblock {\em Communications Earth \& Environment}, 6(1):124, 2025.

\bibitem{molteni1996ecmwf}
Franco Molteni, Roberto Buizza, Tim~N Palmer, and Thomas Petroliagis.
\newblock The ecmwf ensemble prediction system: Methodology and validation.
\newblock {\em Quarterly journal of the royal meteorological society}, 122(529):73--119, 1996.

\bibitem{raffa2021cosmo}
Mario Raffa, Alfredo Reder, Gian~Franco Marras, Marco Mancini, Gabriella Scipione, Monia Santini, and Paola Mercogliano.
\newblock Vhr-rea\_it dataset: Very high resolution dynamical downscaling of era5 reanalysis over italy by cosmo-clm.
\newblock {\em Data}, 6(8), 2021.

\bibitem{RampalRALSD}
Neelesh Rampal, Peter~B. Gibson, Steven Sherwood, Gab Abramowitz, and Sanaa Hobeichi.
\newblock A reliable generative adversarial network approach for climate downscaling and weather generation.
\newblock {\em Journal of Advances in Modeling Earth Systems}, 17(1):e2024MS004668, 2025.
\newblock e2024MS004668 2024MS004668.

\bibitem{rasp2018}
Stephan Rasp and Sebastian Lerch.
\newblock Neural networks for postprocessing ensemble weather forecasts.
\newblock {\em Monthly Weather Review}, 146(11):3885–3900, November 2018.

\bibitem{Ravuri_2021}
Suman Ravuri, Karel Lenc, Matthew Willson, Dmitry Kangin, Remi Lam, Piotr Mirowski, Megan Fitzsimons, Maria Athanassiadou, Sheleem Kashem, Sam Madge, Rachel Prudden, Amol Mandhane, Aidan Clark, Andrew Brock, Karen Simonyan, Raia Hadsell, Niall Robinson, Ellen Clancy, Alberto Arribas, and Shakir Mohamed.
\newblock Skilful precipitation nowcasting using deep generative models of radar.
\newblock {\em Nature}, 597(7878):672–677, September 2021.

\bibitem{ritchie1995implementation}
Harold Ritchie, Clive Temperton, Adrian Simmons, Mariano Hortal, Terry Davies, David Dent, and Mats Hamrud.
\newblock Implementation of the semi-lagrangian method in a high-resolution version of the ecmwf forecast model.
\newblock {\em Monthly Weather Review}, 123(2):489--514, 1995.

\bibitem{rockel2008performance}
Burkhardt Rockel and Beate Geyer.
\newblock The performance of the regional climate model clm in different climate regions, based on the example of precipitation.
\newblock {\em Meteorologische Zeitschrift (Berlin)}, 17, 2008.

\bibitem{rodrigues2018}
Eduardo~R. Rodrigues, Igor Oliveira, Renato L.~F. Cunha, and Marco A.~S. Netto.
\newblock Deepdownscale: a deep learning strategy for high-resolution weather forecast, 2018.

\bibitem{rombach2022high}
Robin Rombach, Andreas Blattmann, Dominik Lorenz, Patrick Esser, and Bj{\"o}rn Ommer.
\newblock High-resolution image synthesis with latent diffusion models.
\newblock In {\em Proceedings of the IEEE/CVF conference on computer vision and pattern recognition}, pages 10684--10695, 2022.

\bibitem{ruhling2023dyffusion}
Salva R{\"u}hling~Cachay, Bo~Zhao, Hailey Joren, and Rose Yu.
\newblock Dyffusion: A dynamics-informed diffusion model for spatiotemporal forecasting.
\newblock {\em Advances in neural information processing systems}, 36:45259--45287, 2023.

\bibitem{saoulis2025diffusion}
Alexandros~Angelos Saoulis, Chris Lucas, Natalie~S. Lord, Nans Addor, and Jorge~Sebastián Moraga.
\newblock Diffusion models for climate data surpass alternative statistical downscaling techniques.
\newblock {\em Authorea}, 2 2025.
\newblock Preprint.

\bibitem{schmidt2024spatiotemporally}
Jonathan Schmidt, Luca Schmidt, Felix Strnad, Nicole Ludwig, and Philipp Hennig.
\newblock Spatiotemporally coherent probabilistic generation of weather from climate.
\newblock {\em arXiv preprint arXiv:2412.15361}, 2024.

\bibitem{sohl2015deep}
Jascha Sohl-Dickstein, Eric Weiss, Niru Maheswaranathan, and Surya Ganguli.
\newblock Deep unsupervised learning using nonequilibrium thermodynamics.
\newblock In {\em International conference on machine learning}, pages 2256--2265. pmlr, 2015.

\bibitem{srivastava2024precipitation}
Prakhar Srivastava, Ruihan Yang, Gavin Kerrigan, Gideon Dresdner, Jeremy McGibbon, Christopher~S Bretherton, and Stephan Mandt.
\newblock Precipitation downscaling with spatiotemporal video diffusion.
\newblock {\em Advances in Neural Information Processing Systems}, 37:56374--56400, 2024.

\bibitem{tomasi2024can}
Elena Tomasi, Gabriele Franch, and Marco Cristoforetti.
\newblock Can ai be enabled to dynamical downscaling? a latent diffusion model to mimic km-scale cosmo5. 0$\backslash$\_clm9 simulations.
\newblock {\em arXiv preprint arXiv:2406.13627}, 2024.

\bibitem{sevir}
Mark Veillette, Siddharth Samsi, and Chris Mattioli.
\newblock Sevir : A storm event imagery dataset for deep learning applications in radar and satellite meteorology.
\newblock In H.~Larochelle, M.~Ranzato, R.~Hadsell, M.F. Balcan, and H.~Lin, editors, {\em Advances in Neural Information Processing Systems}, volume~33, pages 22009--22019. Curran Associates, Inc., 2020.

\bibitem{watt2024generative}
Robbie~A Watt and Laura~A Mansfield.
\newblock Generative diffusion-based downscaling for climate.
\newblock {\em arXiv preprint arXiv:2404.17752}, 2024.

\bibitem{book}
Pieter Wesseling.
\newblock {\em Principles of Computational Fluid Dynamics}.
\newblock Springer Series in Computational Mathematics. Springer Berlin, Heidelberg, 1 edition, 2001.

\bibitem{canmach}
Jonathan~A. Weyn, Dale~R. Durran, and Rich Caruana.
\newblock Can machines learn to predict weather? using deep learning to predict gridded 500-hpa geopotential height from historical weather data.
\newblock {\em Journal of Advances in Modeling Earth Systems}, 11(8):2680--2693, 2019.

\end{thebibliography}

\pagebreak

\appendix
\section{Features Used for Input}
\label{app:features}

As described within our problem formulation in Section \ref{subsec:formulation}, low-resolution data and high-resolution static data are used as model inputs. Our low-resolution input consists of 11 features from the ERA5 reanalysis, namely
\begin{itemize}
    \item 2-m temperature
    \item 10-m zonal and meridional wind speed
    \item mean sea level pressure
    \item sea surface temperature
    \item snow depth
    \item dew-point 2-m temperature
    \item incoming surface solar radiation
    \item temperature at 850 hPa
    \item zonal, meridional and vertical wind speed at 850hPa
    \item specific humidity at 850 hPa
    \item total precipitation
\end{itemize}

Our high-resolution static data consists of three features, which are
\begin{itemize}
    \item topography (from a digital elevation model)
    \item land cover
    \item latitude
\end{itemize}

\section{Additional Visuals of Temperature Downscaling vs Benchmarks}
\label{app:figs}

\begin{figure}[H]
    \centering
    \includegraphics[width=0.9\linewidth]{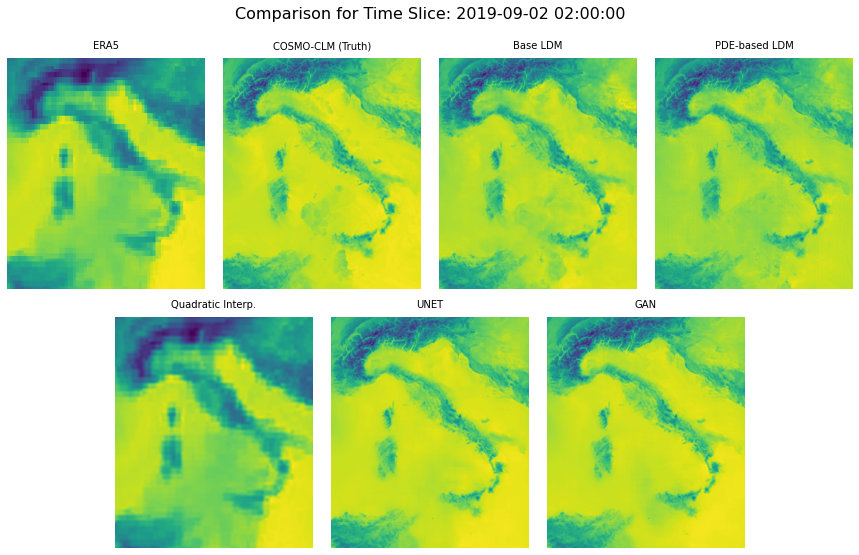}
    \label{fig:comparison1}
\end{figure}

\begin{figure}[H]
    \centering
    \includegraphics[width=0.9\linewidth]{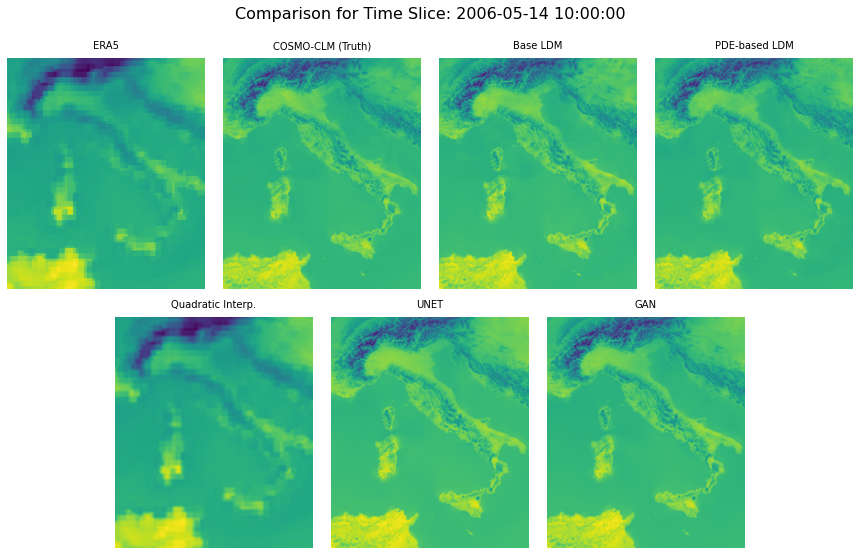}
    \label{fig:comparison2}
    
\end{figure}

\begin{figure}[H]
    \centering
    \includegraphics[width=0.9\linewidth]{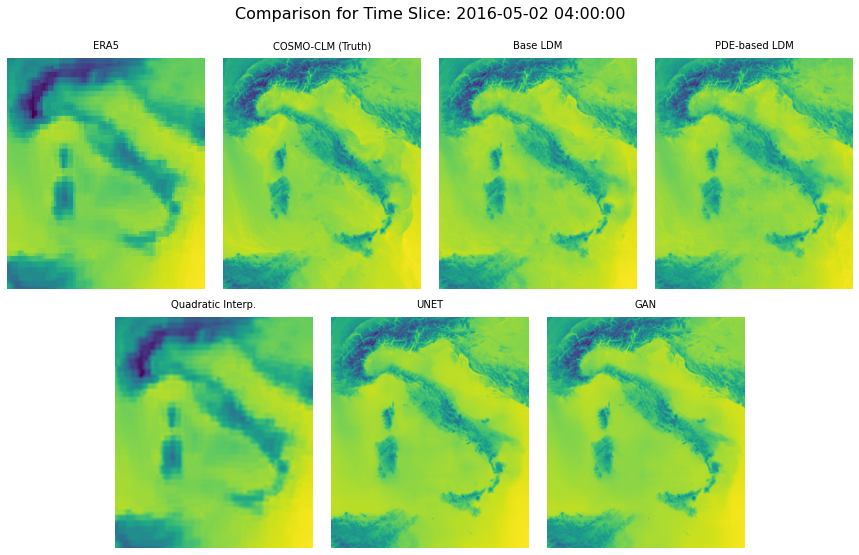}
    \label{fig:comparison4}
\end{figure}

\begin{figure}[H]
    \centering
    \includegraphics[width=0.9\linewidth]{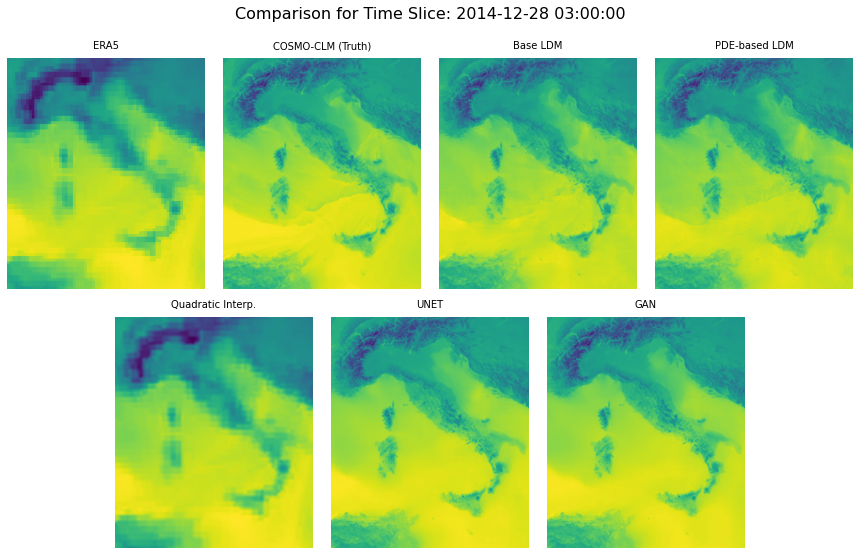}
    \label{fig:comparison5}
\end{figure}

\end{document}